\documentclass{article}

\PassOptionsToPackage{numbers}{natbib}

 \usepackage[preprint]{neurips_2026}

\usepackage[utf8]{inputenc} 
\usepackage[T1]{fontenc}    
\usepackage[hypertexnames=false]{hyperref}       
\usepackage{url}            
\usepackage{booktabs}       
\usepackage{amsfonts}       
\usepackage{nicefrac}       
\usepackage{microtype}      
\usepackage{xcolor}         

\usepackage{amsmath, amssymb}
\usepackage{enumitem}
\usepackage{amsmath}
\usepackage{amsthm}
\usepackage{bm}
\usepackage{thm-restate}
\usepackage{thmtools}
\usepackage{graphicx}
\usepackage{algorithm}
\usepackage{algpseudocode}
\usepackage{multirow}   
\usepackage{siunitx}
\usepackage{hyperref}
\usepackage{placeins}
\usepackage{makecell}
\usepackage{tabularx}
\definecolor{good}{RGB}{0,130,60}
\definecolor{bad}{RGB}{190,30,45}
\definecolor{ourrow}{RGB}{235,242,250}

\newcommand{\up}[1]{{\color{good}\scriptsize$\blacktriangle$\,#1}}
\newcommand{\dn}[1]{{\color{bad}\scriptsize$\blacktriangledown$\,#1}}
\usepackage{colortbl}
\theoremstyle{definition}

\theoremstyle{remark}

\title{RIZZ: Routing Interactions to Near Zero-Interference Zones for Continual Adaptation of Black-Box Agents}

  \author{%
  Sonali Goel$^\dagger$, Pranav Vaidhyanathan$^\dagger$, Lucas Schorling, Natalia Ares, and Maike Osborne \\
  Department of Engineering Science\\
  University of Oxford\\
\texttt{Corresponding Author: pranav@robots.ox.ac.uk}
\thanks{$^\dagger$: Equal Contribution} \\
}

\begin{document}

\maketitle

\begin{abstract}

 Large language models are increasingly deployed as long-lived agents that must adapt across users, tasks, domains, modalities, and feedback regimes without access to model weights. Existing black-box adaptation methods typically optimize a single prompt, maintain an undifferentiated memory, or rely on repeated rollout-heavy search. However, these designs struggle when streams of input are nonstationary, feedback is sparse, and failures from one task family can contaminate behavior on another. We introduce RIZZ (\textbf{R}outing \textbf{I}nteractions to Near \textbf{Z}ero-interference \textbf{Z}ones), a continual adaptation framework for compound language-model systems that learns entirely through verifier-gated memory, routing, and prompt compilation. RIZZ organizes input streams into dynamically spawned memory branches. At inference time, either while online or offline, a context-aware router selects or creates a branch that retrieves branch-local, global, graph-structured, and working-memory context, which is compiled into a bounded prompt together with retrieved task evidence. After the model acts, task verifiers score the output, and only verified interactions can update memory, promote reusable rules, demote harmful rules, or create anti-patterns. This yields a black-box agent that improves through persistent natural-language feedback while explicitly controlling interference. RIZZ targets the regime where adaptation must occur online under context budgets. Finally, we demonstrate the effectiveness of our framework against state-of-the-art baselines on competitive benchmarks.

\end{abstract}

\section{Introduction}
\label{sect:introduction}

Frontier language models are no longer used only as static predictors~\cite{huang2024understanding}. They are increasingly deployed inside compound agents built around tools, retrievers, memory stores, verifiers, routers, prompt programs, and local controllers \cite{compoundaiblog,khattab2023dspycompilingdeclarativelanguage,yao2023reactsynergizingreasoningacting,schick2023toolformerlanguagemodelsteach}. In this setting, the model weights stay fixed, so adaptation has to happen through the surrounding system rather than through fine-tuning. The core challenge is to make the agent improve from experience without changing the base model or contaminating one task with the lessons of another \cite{qiu2024continuallearningusinglarge}. This challenge is especially acute for API-only models, where online parameter updates are unavailable, unsafe, or too expensive \cite{singh2026openaigpt5card, qi2023finetuningalignedlanguagemodels}.

A deployed agent observes a stream
\[
(x_1,\hat y_1, r_1),(x_2,\hat y_2, r_2),\ldots,
\]
where each \(x_t\) is a query drawn from some latent task family, user context, or modality, \(\hat y_t\) is the agent’s response, and each \(r_t\) is the reward revealed only after the agent has produced its response~\cite{goodfellow2013empirical}. The agent must therefore make each prediction before seeing its reward, then decide whether that reward is strong enough to justify updating its external memory state. Continual learning on non-stationary streams is prone to catastrophic forgetting and interference, while task similarity can either support transfer or exacerbate interference depending on how tasks overlap \citep{parisi2019continuallifelonglearning,flesch2023continualtasklearning}. A single global memory is brittle because lessons from one task family can interfere with another, even when the tasks are only partially related \cite{dun2025sweepingheterogeneitysmartmops, zhang2026agenticcontextengineeringevolving}: a rule learned for protein structure prediction may not transfer cleanly to protein function annotation, while a shortcut learned from conversational summarization may add noise to retrieval or tool use. But complete isolation is also wasteful, since online adaptation should compound wherever related structure recurs.

Biological memory offers a useful design intuition. Organisms do not rebuild their neural machinery after every experience. Memory begins plastic, then becomes selective, and eventually stabilizes into durable habits as evidence accumulates \cite{cowan2021memoryconsolidation, tome2024dynamic}. This suggests a simple lifecycle for continual adaptation: absorb new experience when it is still sparse, specialize when patterns recur, and stabilize once useful structure has been validated.

We introduce RIZZ, a lightweight framework for continual online adaptation through routed external memory. For each new query, the system selects a specialized memory zone, or \textit{branch}, shaped by prior cases that required similar behavior. Each branch functions as a small expert memory: it stores successful examples, distills recurring patterns into procedural rules, and tracks whether those rules help or hurt on future calls. When a query falls outside existing experience, a new branch can be created; when two branches become redundant, they can be merged or pruned.

Memory updates are gated by verifier feedback. After the model answers, a deterministic verifier (such as fuzzy ratio) scores the result, and only sufficiently successful interactions are written into durable memory, while weak or harmful traces are demoted, quarantined, or discarded. RIZZ is task-agnostic: it does not require ground-truth task IDs or a pre-specified taxonomy, and instead discovers its own memory organization from query content, routing signals, and verifier rewards. This makes dynamic memory management the core learning problem. At each step, the system chooses which branch should condition the model, which memories fit within the context budget, and whether the outcome is worth writing back. Adaptation therefore happens through routing, retrieval, curation, and prompt compilation, while the underlying language model remains fixed.

RIZZ delivers high accuracy at low token cost. Using Claude Haiku 4.5~\cite{anthropic2025haiku45} as the frozen base model (in the following denoted as method Frozen), it consistently improves over the no-memory baseline across StreamBench~\cite{wu2024streambenchbenchmarkingcontinuousimprovement}, TRACE~\cite{wang2023tracecomprehensivebenchmarkcontinual}, LongMemEval-S~\cite{wu2025longmemevalbenchmarkingchatassistants}, and $\tau$-Bench~\cite{yao2024taubenchbenchmarktoolagentuserinteraction}, while staying close to Frozen in cost and far cheaper than heavier memory baselines. Because adaptation happens through routing and memory updates rather than fine-tuning, training time remains low as well. The gains appear where memory should matter most: recurring task structure, session-local conversational evidence, and repeated domain policies in tool-mediated interaction. The result is a reliable online adaptation framework that is accurate, efficient, and practical to deploy.

Our contributions are as follows:
\begin{itemize}[leftmargin=*]
    \item \textbf{A task-agnostic cost and token efficient routing framework.} We formulate continual black-box adaptation as an online decision problem in which a frozen model adapts only through external state. RIZZ requires no ground-truth task IDs or pre-specified task taxonomy, and it discovers specialist memory branches online from query content and verifier feedback while being token and cost efficient.

    \item \textbf{Compact branch-local memory with hierarchical routing.} We introduce routed branch memory together with a hierarchical routing mechanism that combines output-shape guards, function and application labels, embedding-based snapping, and Jaccard de-duplication. This keeps related experience together, limits cross-task interference, and allows branches to be created, reused, and merged without supervised task labels.

    \item \textbf{Verifier-gated memory evolution under tight budgets.} We make memory updates depend on reward validation: successful examples enter memory, procedural rules are reweighted by helpful and harmful use, failures are retained as concrete anti-templates, and redundant branches are pruned or merged. A budget-aware prompt compiler then assembles branch-local evidence under a fixed context window, so the system degrades gracefully to frozen rather than inflating token cost.

    \item \textbf{Strong, lightweight empirical results.} Across streaming adaptation, sequential continual learning, long-term conversational memory, and tool-use benchmarks, RIZZ consistently improves over Frozen while staying cost-competitive and substantially cheaper than heavier memory baselines. In particular, it delivers high accuracy without fine-tuning, rollout-heavy search, or unbounded context growth.
\end{itemize}
\section{Related Work}

RIZZ lies at the intersection of black-box language-model adaptation, reflective prompt optimization, test-time memory, agentic context engineering, and long-term memory for LLM agents. In all of these areas, systems improve without updating model parameters; the critical distinction is \emph{what object is adapted}. Reflexion~\cite{shinn2023reflexionlanguageagentsverbal} and Self-Refine~\cite{selfrefine} established verbal self-improvement through reflection and iterative critique. TextGrad~\cite{yuksekgonul2024textgradautomaticdifferentiationtext}, GEPA~\cite{GEPAmain}, and Feedback Descent~\cite{lee2025feedbackdescentopenendedtext} generalized this into frameworks for optimizing prompts and text artifacts via LLM-generated feedback. Dynamic Cheatsheet~\cite{suzgun2025dynamiccheatsheettesttimelearning} and ACE~\cite{zhang2026agenticcontextengineeringevolving} gave agents persistent, evolving memory and playbooks at inference time. RAG~\cite{lewis2021retrievalaugmentedgenerationknowledgeintensivenlp}, MemGPT~\cite{packer2024memgptllmsoperatingsystems}, and A-Mem~\cite{xu2025amemagenticmemoryllm} developed increasingly sophisticated retrieval, tiered context management, and self-organizing memory structures. Benchmarks such as StreamBench~\cite{wu2024streambenchbenchmarkingcontinuousimprovement}, TRACE~\cite{wang2023tracecomprehensivebenchmarkcontinual}, and LongMemEval~\cite{wu2025longmemevalbenchmarkingchatassistants} measure whether such adaptation persists without catastrophic interference.

RIZZ also connects to continual-learning work in parameter space, where the central concern is preserving plasticity while limiting interference as tasks arrive sequentially. Representative examples include Maintaining Plasticity in Deep Continual Learning, which studies loss of plasticity and proposes continual backpropagation~\citep{dohare2023maintainingplasticity}; ELLA, a lightweight adapter-based lifelong learning method with constant memory and compute~\citep{biswas2026ella}; TRC$^2$, which uses sparse thalamic routing over cortical columns for continual adaptation~\citep{khadangi2026trc2}; and ARES, which alternates reinforcement learning and supervised fine-tuning using teacher feedback~\citep{byun2024ares}. These methods target weight-space adaptation rather than external memory, but they share the same goal of improving a model over a stream of tasks while controlling interference.

RIZZ differs from these systems in three respects. First, where prior methods maintain a single prompt, context, or global memory, RIZZ decomposes state into dynamically spawned branches, each with procedural rules, episodic exemplars~\cite{lopezpaz2017gems}, and reward statistics. Second, all memory writes are gated on task-verifier feedback rather than self-curation. Third, routing each query to the right branch is treated as a first-class learning problem. The following sections detail this design, and a comprehensive discussion of related work is provided in Appendix~\ref{app:relatedwork}.

\section{Methods}
\label{sec:methods}

\begin{figure*}[h]
\centering
\includegraphics[width=\textwidth]{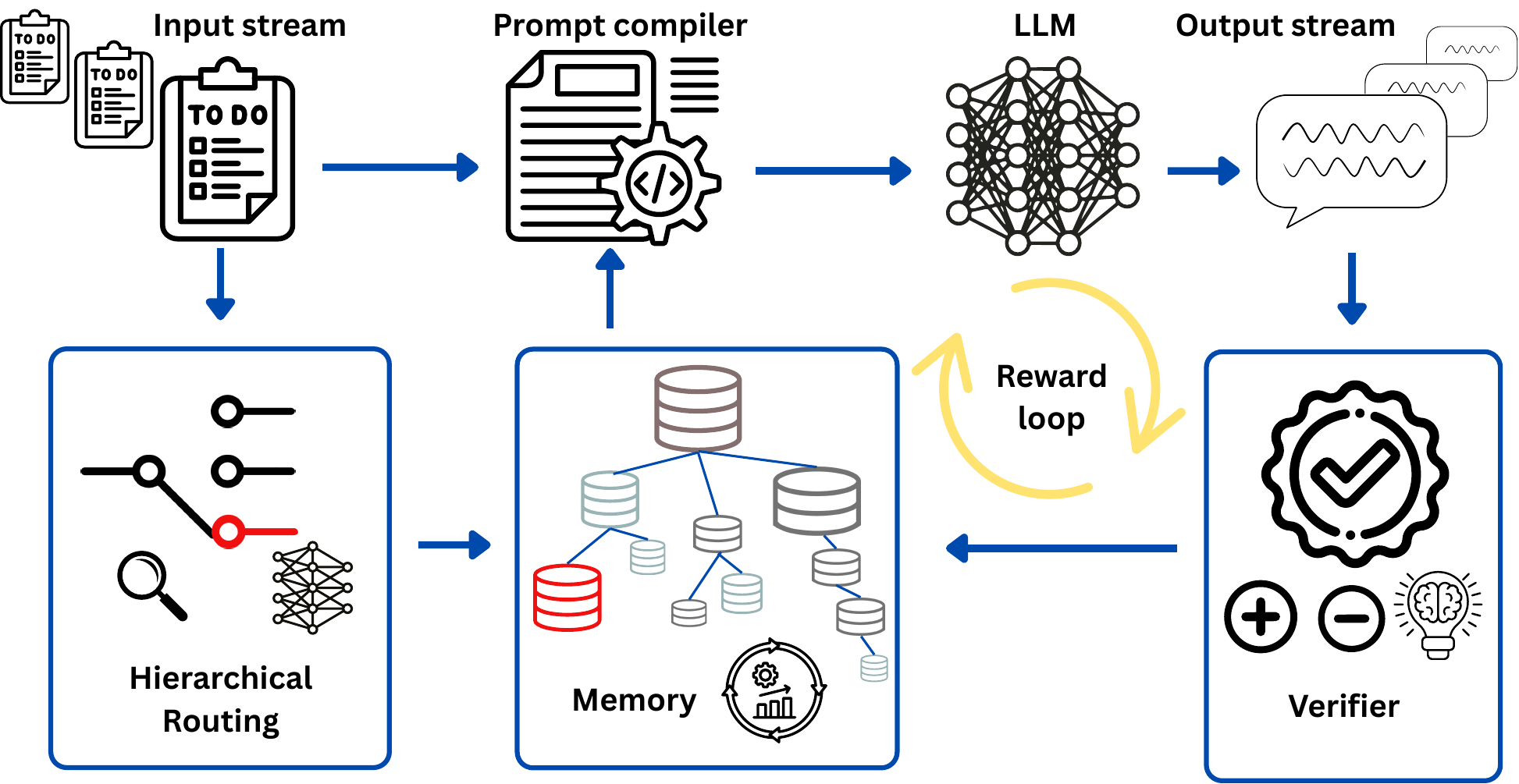}
\caption{\textbf{Simplified schematic of RIZZ.} RIZZ adapts a frozen black-box language model through external routed memory rather than parameter updates. Each incoming input is assigned to a separated branch by lightweight routing. The selected branch supplies episodic examples, procedural rules, common mistakes, and routing statistics to a bounded prompt compiler. A task verifier scores the model output; only verifier-gated feedback can update memory, credit or debit rules, and spawn, merge, or prune branches.
}
\label{fig:schematic}
\end{figure*}

RIZZ is designed around a simple biological intuition: adaptation does not require repeatedly rewiring the core machinery, but rather keeping what matters, pruning what misleads, and consolidating steady experience into durable memory \cite{cowan2021memoryconsolidation,frey1997synaptic,tononi2014sleep}. We translate this idea to frozen black-box language models by shifting adaptation out of the parameters and into an external memory that evolves \textit{online} as new task families appear. Rather than storing all experience in one global memory, RIZZ routes each interaction into a \emph{zero-interference zone}: a separated branch that accumulates local examples, failures, context, and reward statistics. At inference time either offline or online, RIZZ routes the input, retrieves branch-conditioned memory under a fixed context budget, calls the frozen model, verifies the output, and writes back only when feedback justifies doing so. The base model is never updated. A simplified schematic is shown in Figure \ref{fig:schematic}.

\begin{algorithm}[h]
\caption{RIZZ online step}
\label{alg:rizz-online}
\small
\begin{algorithmic}[1]
\Require Frozen model $M$; embedder $\phi$; verifier $\mu_t$; state $S_t=(\mathcal{B}_t,\mathcal{W}_t,\mathcal{H}_t)$; input $z_t=(x_t,I_t,m_t,d_t,y_t^\star)$
\Ensure Output $\hat y_t$ for inputs; updated state $S_{t+1}$
\State $e_t \gets \phi(x_t)/\|\phi(x_t)\|_2$
\State $(b_t,\mathrm{mode}) \gets \operatorname{Route}(z_t,e_t,S_t)$
\If{$\mathrm{mode}=\textsc{store}$}
    \State $S_{t+1}\gets \operatorname{Update}(S_t,b_t,z_t,e_t)$
    \State \Return $S_{t+1}$
\EndIf
\State $\Gamma_t \gets \operatorname{Retrieve}(x_t,e_t,b_t,S_t,m_t,d_t)$
\State $p_t \gets \operatorname{Compile}(x_t,I_t,\Gamma_t,B_{\rm ctx})$
\State $\tilde y_t \gets M(p_t)$
\State $\hat y_t \gets \operatorname{FinalAnswer}(\tilde y_t)$
\State $r_t \gets \mu_t(\hat y_t,y_t^\star,x_t,m_t)$
\If{$\operatorname{Writable}(m_t)$}
    \State $S_{t+1}\gets \operatorname{Update}(S_t,b_t,z_t,\hat y_t,r_t,e_t,\Gamma_t)$
\Else
    \State $S_{t+1}\gets S_t$
\EndIf
\State \Return $(\hat y_t,r_t,S_{t+1})$
\end{algorithmic}
\end{algorithm}

As seen in Algorithm \ref{alg:rizz-online}, at step $t$, RIZZ receives an input stream $z_t=(x_t,I_t,m_t,d_t,y_t^\star)$, where $x_t$ is either an action query or context, $I_t$ is an optional instruction, $m_t$ is metadata, $d_t$ is optional external evidence, and $y_t^\star$ is used only by the verifier. For the input stream, a frozen model $M$ is prompted to produce an answer $\hat y_t$, and a verifier $\mu_t$ returns a reward $r_t\in[0,1]$. All routing and retrieval use a frozen normalized embedder $\phi$, implemented in our runs with Qwen3-VL embeddings \cite{li2026qwen3vlembeddingqwen3vlrerankerunifiedframework}. 

\subsection{Adaptive state and routing}
The external state is provided $S_t=(\mathcal{B}_t,\mathcal{W}_t,\mathcal{H}_t)$: a branch bank, bounded working memory, and routing/reward diagnostics. Each branch $b\in\mathcal{B}_t$ stores labels, successful episodic examples as well as verifier-weighted procedural rules, substantive failures, scoped context, and summary statistics.

RIZZ routes before retrieval. The router asks an LLM judge for a hierarchical label, composed of the abstracted branch function $f$ and the application domain $a$ . Near-duplicate labels are snapped to existing labels in embedding space, and same-function sibling branches are de-duplicated by token overlap in the application slot. Concretely, for a normalized label $\ell_t=(f_t,a_t)$, RIZZ snaps to an existing sibling $b$ when $\cos(\psi(a_t),\psi(a_b))\geq\tau_{\rm snap}=0.92$; otherwise, among branches with $f_b=f_t$, it reuses $b$ when $J(a_t,a_b)=|\mathcal{T}(a_t)\cap\mathcal{T}(a_b)|/|\mathcal{T}(a_t)\cup\mathcal{T}(a_b)|\geq\tau_J=0.40$, where $\psi$ embeds label strings and $\mathcal{T}$ tokenizes application labels (these hyperparameters are often on a stable plateau as seen in Table ~\ref{tab:threshold_sensitivity}). As an illustration, a query asking for a Python code snippet would first receive the shape tag \texttt{code::latin}, and then a judge-assigned label for this might look like \texttt{complete-code-snippet/numpy-array}. An existing branch with that exact label is preferred to a sibling matching only the function (e.g.\ \texttt{complete-code-snippet/pandas-dataframe}) or only the application. If no compatible branch exists and capacity remains, RIZZ spawns a new branch. For no questions to label, RIZZ routes by nearest spawn anchor and spawns a new context branch when the input is sufficiently novel. Formal routing scores and thresholds are given in Appendix~\ref{app:rizz-formalism}.

\subsection{Branch bank and memory}

Each branch contains multiple active memory substrates. Episodic memory stores verifier-approved input/output examples \cite{lopezpaz2017gems}. Procedural memory stores compact IF--THEN rules distilled from batches of successful examples \cite{zhang2026agenticcontextengineeringevolving}. Each rule statistic maintains helpful and harmful counters that affect future retrieval. Failure memory stores substantive wrong answers as concrete ``common mistakes'' rather than broad negative rules. RIZZ decouples read and write locality: routing selects the branch that will receive updates, but positive examples and rules may be retrieved across branches, while failure memory remains local to avoid overgeneralizing mistakes. Session-like context is first filtered by metadata scope when available, then retrieved by a mixture of semantic similarity and recency.

\subsection{Prompt compilation}
The compiler converts retrieved material $\Gamma_t$ into a bounded prompt. It uses a fixed priority cascade: scoped context, episodic examples, common mistakes, recent working memory, and finally the query. Lower-priority blocks are dropped first when the token budget is tight, so RIZZ degrades smoothly toward frozen prompting rather than exceeding the context window. The prompt requests a stable final-answer field; the verifier is applied to the extracted answer rather than to the full model output. 

\subsection{Verifier-gated memory evolution.}
Writes occur when online learning is enabled. Retrieved rules receive helpful or harmful credit from the verifier (see Appendix \ref{app:ver} for more details). Successful interactions enter episodic memory if they clear the verifier-dependent write threshold; substantive failures may enter the local failure store. A quarantine guard prevents low-reward, off-centroid interactions from contaminating the selected branch. Periodic lifecycle sweeps merge redundant branches and prune stale low-utility ones, keeping the routing space compact.

\section{Evaluation}

We evaluate RIZZ on four benchmarks that stress complementary aspects of continual black-box adaptation. TRACE \cite{wang2023tracecomprehensivebenchmarkcontinual} tests sequential continual learning across heterogeneous task families, including multilingual understanding, code, math, summarization, and domain-specific reasoning. StreamBench \cite{wu2024streambenchbenchmarkingcontinuousimprovement} evaluates online improvement from streams of input-feedback pairs without weight updates. LongMemEval-S \cite{wu2025longmemevalbenchmarkingchatassistants} probes long-term conversational memory across sessions, including recall, temporal reasoning, and preference tracking, and $\tau$-Bench \cite{yao2024taubenchbenchmarktoolagentuserinteraction} evaluates tool-using agents in realistic retail and airline conversations where success requires policy-compliant API use.  We compare against ACE (Agentic Context Engineering) \cite{zhang2026agenticcontextengineeringevolving}, which updates an evolving playbook through generation, reflection, and curation, and AMem (Agentic Memory) \cite{xu2025amemagenticmemoryllm}, which organizes memory as a dynamic, Zettelkasten-inspired network of linked notes. 

\begin{table}[h]
\centering
\caption{\textbf{Main results} on continual-learning benchmarks (Claude Haiku 4.5).
Cost computed at Haiku 4.5 rates (\$1\,/\,M input + \$5\,/\,M output).
Deltas are percentage-point changes in average reward vs.\ the Frozen baseline.
\textsuperscript{*}\,context-window collapse.}
\label{tab:main_results}
\renewcommand{\arraystretch}{1.2}
\setlength{\tabcolsep}{8pt}
\small
\begin{tabular}{@{}ll r@{\hspace{1em}}l rrr@{}}
\toprule
\textbf{Benchmark} & \textbf{Method} &
  \multicolumn{2}{c}{\textbf{Avg. reward}} &
  \textbf{Cost (\$)} & \textbf{Tokens (M)} & \textbf{Wall (h)} \\
\midrule
\multirow{4}{*}{\textsc{StreamBench}}
  & Frozen                 & 0.6013 &                & 7.99   & 5.13   & 2.68  \\
  & AMem                   & 0.6234 & \up{2.2 pp}       & 57.75  & 36.87  & 27.40 \\
  & ACE\textsuperscript{*} & 0.1720 & \dn{42.93 pp}      & 292.23 & 283.53 & 13.07 \\
\rowcolor{ourrow}
  & \textbf{RIZZ}\,\emph{(ours)} & \textbf{0.6339} & \up{3.3 pp} & 7.37 & 5.14 & 7.23 \\
\midrule
\multirow{4}{*}{\textsc{TRACE}}
  & Frozen                 & 0.6369 &                & 7.51  & 3.95  & 2.86 \\
  & AMem                   & 0.6302 & \dn{0.7 pp}       & 18.11 & 13.97 & 4.05 \\
  & ACE\textsuperscript{*} & 0.0472 & \dn{59.0 pp}      & 46.52 & 43.36 & 3.04 \\
\rowcolor{ourrow}
  & \textbf{RIZZ}\,\emph{(ours)} & \textbf{0.7060} & \up{6.9 pp} & 12.79 & 7.92 & 7.99 \\
\midrule
\multirow{4}{*}{\textsc{LongMemEval-S}}
  & Frozen                 & 0.2020 &                & 6.80  & 9.03  & 0.43  \\
  & AMem                   & 0.3220 & \up{12.0 pp}      & 128.10   & 82.9  & 36.80   \\
  & ACE\textsuperscript{*}                    & 0.082 & \dn{12.0 pp}      & 30.29 & 31.16 & 12.50 \\
\rowcolor{ourrow}
  & \textbf{RIZZ}\,\emph{(ours)} & \textbf{0.4836}     & \up{28.2 pp} & 12.43 & 12.17 & 18.16 \\
\bottomrule
\end{tabular}
\end{table}

Across these settings (Tables~\ref{tab:main_results}, \ref{tab:tau_bench}, and in Appendix \ref{app:per-sub-task} Table \ref{tab:per_subtask}), RIZZ is the only framework that reliably improves over the no-memory \emph{Frozen} baseline while staying operationally lightweight. The pattern is consistent: RIZZ wins when streams contain reusable local structure, such as repeated output formats, recurring solution templates, session-specific evidence, or domain-specific tool policies. Rather than storing all experience in one global memory, RIZZ routes related interactions into compact branches where useful procedures accumulate without contaminating unrelated tasks.

\newcommand{\ov}[2]{\makebox[0.78cm][r]{#1}\hspace{0.25em}\makebox[0.95cm][l]{#2}}


\begin{table}[h]
\centering
\caption{\textbf{$\tau$-Bench results} using a ReAct agent on Claude Haiku 4.5.
Overall scores report benchmark pass@1. Costs are computed from total token usage across the evaluation stream.
Deltas are percentage point changes in Overall vs.\ the Frozen baseline. \textbf{Bold} = best in column.}
\label{tab:tau_bench}
\renewcommand{\arraystretch}{1.2}
\setlength{\tabcolsep}{5pt}
\small
\resizebox{\linewidth}{!}{%
\begin{tabular}{@{}lcc r@{\hspace{0.7em}}l rrrr@{}}
\toprule
\textbf{Method} &
  \textbf{Retail} & \textbf{Airline} &
  \multicolumn{2}{c}{\textbf{Overall}} &
  \textbf{Turns} & \textbf{Cost (\$)} & \textbf{Tokens (M)} & \textbf{Wall (h)} \\
\midrule
ReAct + Frozen & 0.357 & 0.340 & 0.352 & \makebox[4.2em][l]{\phantom{\up{4.8 pp}}} & 2194 & 14.57 & 11.75 & 3.0 \\
ReAct + AMem   & 0.304 & 0.300 & 0.303 & \makebox[4.2em][l]{\dn{4.9 pp}}            & \textbf{2061} & \textbf{13.77} & \textbf{11.08} & 3.5 \\
ReAct + ACE    & 0.383 & 0.340 & 0.370 & \makebox[4.2em][l]{\up{1.8 pp}}             & 2260 & 18.53 & 14.76 & 5.8 \\
\rowcolor{ourrow}
\textbf{ReAct + RIZZ}\,\emph{(ours)}
               & \textbf{0.417} & \textbf{0.360} & \textbf{0.400} & \makebox[4.2em][l]{\up{4.8 pp}} & 2177 & 14.48 & 11.64 & 3.5 \\
\bottomrule
\end{tabular}%
}
\end{table}

This selectivity is also what keeps RIZZ cost efficient. The compiler enforces a strict 64K-token cap, playbooks typically stabilize around 10 to 15K tokens, and branch counts remain well below configured limits. ACE exhibits the opposite failure mode: its playbook grows beyond 130K tokens during StreamBench and eventually exceeds Haiku's 200K context window on TRACE, producing hard failures on long-input tasks (see Appendix). AMem avoids context collapse, but its flat, ever-growing memory store gives it substantially higher retrieval cost. RIZZ therefore improves through concentrated memory, not more memory.

\begin{figure*}[h]
\centering
\makebox[\textwidth][c]{%
        \includegraphics[width=\textwidth]{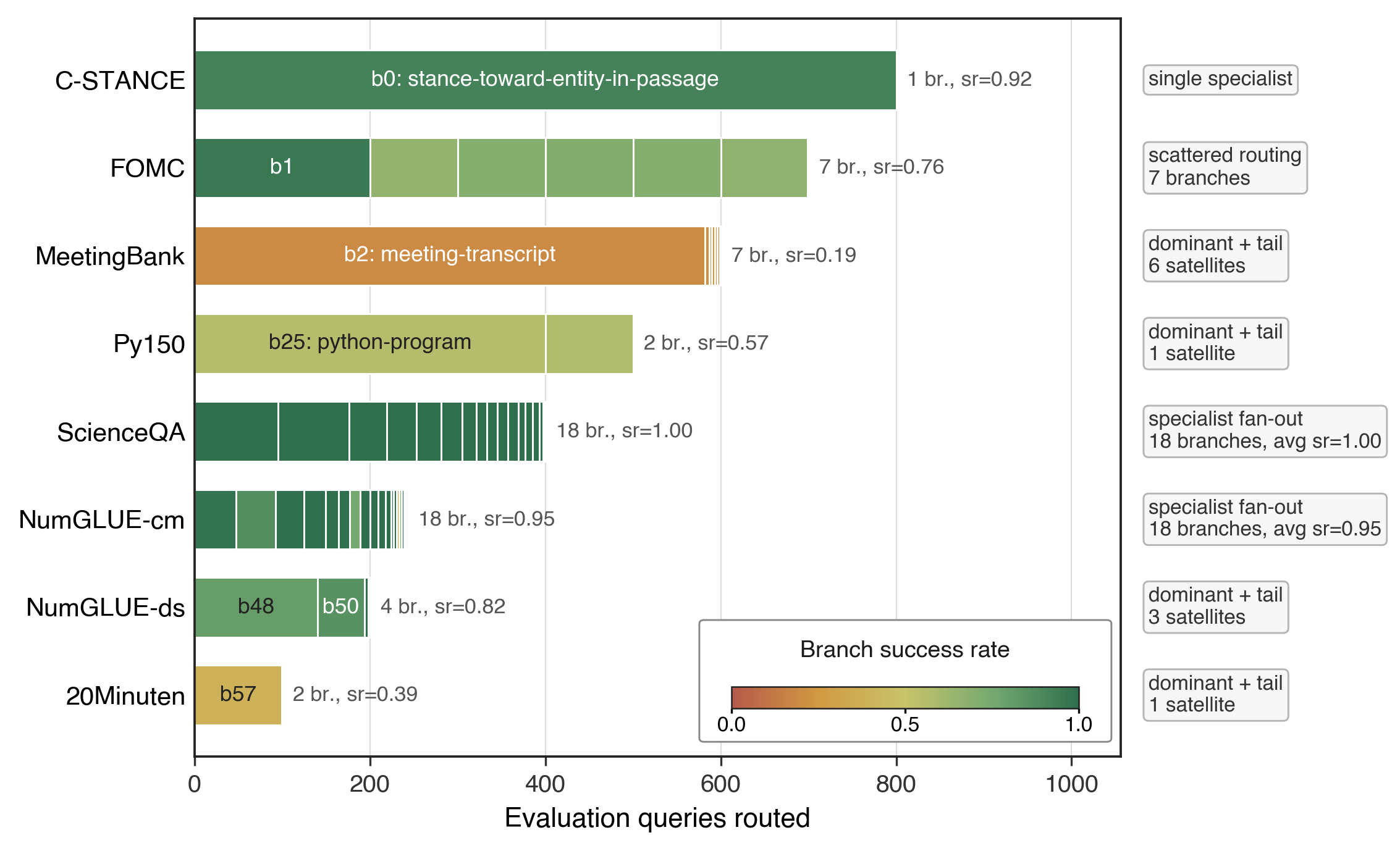}%
    }
\caption{\textbf{Branch routing on TRACE}. Each row is
one TRACE sub-task; segments are branches that received routes from
that sub-task, sized by query count and coloured by branch success
rate. Routing patterns vary systematically with task structure:
sub-tasks with intrinsic compositional structure (\textsc{ScienceQA},
\textsc{NumGLUE-cm}) fan out into many specialist branches at near-
perfect accuracy; sub-tasks with a single coherent task signature
(\textsc{C-STANCE}) collapse into a single high-volume specialist;
sub-tasks without exploitable structure (\textsc{MeetingBank},
\textsc{20Minuten}) accumulate into a dominant branch that fails to
specialize. The pattern is the empirical signature of when RIZZ's
routing finds reusable memory and when it does not.}
\label{fig:trace_branches}
\end{figure*}

\paragraph{TRACE: structured continual learning without playbook collapse.}

TRACE is where RIZZ's branch specialization is most visible (Figure~\ref{fig:trace_branches}). We use the online train/eval split, updating memory on the training portion and freezing it during held-out evaluation. Average reward improves by $+6.9$ points over Frozen and $+7.6$ over AMem ($0.706$ vs.\ $0.637$ / $0.630$), while ACE collapses to $0.047$ after exhausting the context window on long-input tasks. RIZZ beats every baseline on all TRACE subtasks except \textsc{Py150}, \textsc{20Minuten}, where methods are effectively tied, and \textsc{ScienceQA}, where all methods saturate (except for ACE due to the overflow).

The largest gains occur on tasks with repeatable procedural structure: \textsc{C-STANCE} ($+16.4$ points), \textsc{NumGLUE-ds} ($+13.5$), \textsc{FOMC} ($+7.4$), and \textsc{NumGLUE-cm} ($+4.5$). These tasks repeatedly route into coherent branches whose rules sharpen over time. The routing dynamics in Figure~\ref{fig:trace_branches} make the mechanism visible. \textsc{C-STANCE} forms the cleanest case: all $800$ examples route to a single high-success specialist, indicating one stable procedural signature. By contrast, \textsc{ScienceQA} and \textsc{NumGLUE-cm} show specialist fan-out, with many small high-success branches corresponding to narrower sub-skills. \textsc{MeetingBank}, \textsc{Py150}, \textsc{NumGLUE-ds}, and \textsc{20Minuten} follow a dominant-plus-tail pattern, where one main branch absorbs most traffic and a few satellites capture residual variation. \textsc{FOMC} is more diffuse: its policy-statement cues scatter across several branches, but still yield a net gain, suggesting that RIZZ can extract useful signal even when the routing geometry is less sharply separated. Importantly, we observe no cross-contamination across sub-datasets: queries from one sub-dataset remain confined to their assigned branches, even though routing is fully task-agnostic and never uses ground-truth task IDs.

The full TRACE run stabilizes at only $48$ branches ($31$ active), while costing less than both ACE and AMem.

\paragraph{LongMemEval-S: session-local memory for conversational recall.}

LongMemEval-S gives RIZZ its strongest absolute gains. Across the four reported question types (\emph{single-session-user}, \emph{single-session-preference}, \emph{multi-session}, and \emph{temporal-reasoning}), RIZZ reaches average reward $0.484$, ahead of AMem ($0.322$), Frozen ($0.202$), and ACE ($0.082$). The gains concentrate where persistent conversational memory matters most: \emph{single-session-user} recall rises from $0.343 \rightarrow 0.743$, \emph{multi-session} from $0.158 \rightarrow 0.474$, and \emph{temporal-reasoning} from $0.135 \rightarrow 0.346$.

The one exception is \emph{single-session-preference}, where AMem remains strongest ($0.767$ vs.\ RIZZ's $0.533$). This subtype is dominated by short, semantically repetitive preference statements, which a small global similarity store can retrieve effectively. The harder LongMemEval-S cases require session-local evidence to remain available when later questions depend on earlier turns. RIZZ matches this structure directly: across the $357$ evaluated questions, it allocates $195$ branches, effectively treating each session as its own memory locality.

ACE fails for the complementary reason. Its global playbook cannot represent hundreds of disjoint conversational contexts at once, and on LongMemEval-S it eventually saturates Haiku's 200K-token context window. After roughly the first $170$ questions, $320$ of the remaining $326$ questions return \texttt{ContextWindowExceededError}. Its headline reward of $0.082$ therefore mixes two effects: on attempted questions in the reported subset ($n{=}179$), ACE scores $0.168$, while for the rest the memory mechanism prevents generation entirely.

\paragraph{StreamBench: sustained gains under interleaved streams.}

We evaluate StreamBench in a round-robin task-interleaved regime, a deliberately demanding setting for memory systems: related examples are separated by intervening tasks, so useful experience must persist across switches rather than accumulate in clean task blocks. Across $4{,}264$ instances, the controller spawns $83$ branches, merges $3$, prunes $1$, and converges to $79$ active branches, each serving at least one query. Routing nevertheless concentrates around repeated procedural structure: the dominant \texttt{diagnose-from-evidence/clinical-symptoms} branch absorbs $1{,}839$ queries ($43\%$ of traffic), and the top three branches account for $67.7\%$.

The branch population remains stable under this interleaved stream. The drift monitor fires $55$ times, but no branch is rejected for incoherence, suggesting that label-guided routing preserves local consistency even as examples arrive between task switches. RIZZ's strongest StreamBench gains appear where examples repeatedly return to high-volume specialists, especially diagnosis and retrieval-heavy subtasks. Despite this controller activity, RIZZ runs slightly cheaper than the Frozen baseline (\$7.57 vs.\ \$7.99), since the compiled context yields tighter, shorter outputs whose token savings more than offset the modest input-token overhead from the playbook prefix.

DS-1000 exposes the limit of this regime. Its queries fragment across roughly forty low-volume code branches, most with fewer than ten examples, leaving too little repeated experience for procedural or episodic memory to compound. It is the only StreamBench subtask where RIZZ underperforms Frozen. A DS-1000-specific branch-collapsing heuristic, which merged sparse code-completion branches by library family (NumPy, Pandas, PyTorch, SciPy), did not recover performance. The regression therefore appears to reflect a deeper mismatch between sparse library-specific code tasks and the recurrence RIZZ needs for stable branch specialization.

\paragraph{$\tau$-Bench: better tool use at flat cost.}

On $\tau$-Bench, RIZZ improves a ReAct agent without materially increasing cost. Overall pass@1 rises from $0.352$ for Frozen to $0.400$ for RIZZ, ahead of ACE ($0.370$) and AMem ($0.303$). The gain is strongest in retail ($0.417$ vs.\ $0.357$ Frozen), with a smaller but positive gain in airline ($0.360$ vs.\ $0.340$). This setting differs from TRACE and LongMemEval-S: success depends less on long recall and more on repeatedly applying domain policies during tool-mediated conversations. RIZZ remains useful because branch-local memory captures those recurring policy patterns while adding almost no cost relative to Frozen.

\section{Ablations}



Table~\ref{tab:component_ablations} reports the ablation profile for RIZZ on a gpt-4o-mini backbone. Even on a smaller TRACE subset and with a lightweight model, the pattern is clear: RIZZ does not gain by simply retrieving more context, but by accumulating compact, verifier-approved procedures that can be reused within a branch. The branch structure itself is also important. Forcing all traffic through a single branch, or disabling anti-fragmentation and Jaccard sibling snapping, both reduce performance. 

We also report cross-model results for LongMemEval-S in Table~\ref{tab:longmemeval_cross_model}, spanning three frozen backbones from small to large. RIZZ improves consistently across all three. Additional results on multi-seed robustness, hyperparameter sensitivity, and per-task accuracy are provided in the Appendix.

\begin{table}[h]
\centering
\caption{\textbf{Component ablations on TRACE.}
All variants are run on the same 1000-record TRACE-100 uniform sample using \texttt{gpt-4o-mini}.
Each row disables one component of the full RIZZ architecture.
Inline markers show the absolute percentage-point drop relative to full RIZZ on the matched prefix.}
\label{tab:component_ablations}
\renewcommand{\arraystretch}{1.15}
\setlength{\tabcolsep}{7pt}
\small
\begin{tabular}{@{}ll r@{\,\,}l@{}}
\toprule
\textbf{Module} & \textbf{Variant} & \multicolumn{2}{c}{\textbf{Reward}} \\
\midrule
\rowcolor{ourrow}
Full system
& \textbf{RIZZ}\,\emph{(ours)}
& \textbf{0.5051}
& \phantom{{\scriptsize\dn{6.6 pp}}} \\
\midrule
\multirow{1}{*}{Procedural memory}
& No verified memory rendering
& 0.4391
& {\scriptsize\dn{6.6 pp}} \\
\midrule
\multirow{3}{*}{Routing / separation}
& Single branch
& 0.4853
& {\scriptsize\dn{2.0 pp}} \\
& No judge-label anti-fragmentation
& 0.4924
& {\scriptsize\dn{1.3 pp}} \\
& No Jaccard sibling snapping
& 0.4941
& {\scriptsize\dn{1.1 pp}} \\
\bottomrule
\end{tabular}
\end{table}

\begin{table}[tbh!]
\centering
\caption{\textbf{LongMemEval-S cross-model results} comparing RIZZ to Frozen across three frozen backbones of varying sizes. Scores report average reward on the $n{=}100$ subset shown in the figure. Percentage-point improvements are shown inline.}
\label{tab:longmemeval_cross_model}
\small
\renewcommand{\arraystretch}{1.15}
\setlength{\tabcolsep}{8pt}
\begin{tabular}{@{}l r@{\hspace{1.2em}}l r@{\hspace{1.2em}}l r@{\hspace{1.2em}}l@{}}
\toprule
\textbf{Method}
& \multicolumn{2}{c}{\textbf{gpt-4o-mini}}
& \multicolumn{2}{c}{\textbf{Haiku-4.5}}
& \multicolumn{2}{c}{\textbf{gpt-5.1}} \\
\midrule
Frozen
& 0.260 & \phantom{\up{20.0}}
& 0.270 & \phantom{\up{40.0}}
& 0.290 & \phantom{\up{41.0}} \\
\rowcolor{ourrow}
\textbf{RIZZ}\,\emph{(ours)}
& \textbf{0.460} & \up{20.0 pp}
& \textbf{0.670} & \up{40.0 pp}
& \textbf{0.700} & \up{41.0 pp} \\
\bottomrule
\end{tabular}
\end{table}

\section{Conclusion}

We introduced RIZZ, a continual adaptation framework for frozen black-box language-model agents. RIZZ shifts the locus of learning from model parameters to the surrounding compound system: a router assigns each input to a specialized memory branch, a bounded compiler exposes only the most relevant branch-local and cross-branch evidence, and verifier feedback determines which interactions can be written, distilled into reusable procedures, retained as concrete anti-patterns, or discarded. This design connects two previously distinct lines of work. From black-box language feedback and prompt/context optimization, RIZZ inherits the premise that natural language is an effective adaptation substrate. From continual learning, it inherits the central stability--plasticity problem: useful experience should transfer when structure recurs, but incompatible experience should not corrupt behavior on unrelated tasks~\cite{dohare2023maintainingplasticity, GEPAmain}. RIZZ resolves this tension at the memory level rather than the weight level, treating branch creation, branch selection, retrieval, and verifier-gated writeback as the primary adaptation mechanisms.

Across streaming continual learning, heterogeneous sequential tasks, long-term conversational memory, and tool-mediated interaction, RIZZ demonstrates strong performance while remaining operationally lightweight. The empirical pattern is informative. Gains are largest when the stream contains reusable local structure: repeated output formats, recurring procedural templates, session-specific evidence, or domain policies that must be applied repeatedly. In these regimes, branch-local memory lets the agent accumulate compact rules and examples without allowing one task family to dominate a global memory~\cite{wang2022learning}.

\subsection{Limitations and Future Work}
RIZZ improves most when streams contain recurring local structure; when examples are sparse, fragmented, or procedurally heterogeneous, branch-local memory may not accumulate enough evidence to compound, leaving too few verified examples per branch for stable procedural or episodic memory to emerge. Since the base model is frozen, RIZZ can only steer capabilities that the underlying model can express through prompting, retrieved examples, procedural rules, and tool context. Finally, persistent memory introduces privacy, security, fairness, and governance risks. Branch-local storage can reduce cross-task interference, but it is not itself a privacy mechanism: durable memories may contain sensitive user data, or adversarially injected instructions. Practical deployments therefore require retention limits, deletion and correction mechanisms, encryption, access control, provenance metadata, memory auditing, user-visible controls, subgroup evaluation, and defenses against memory poisoning or extraction.

Future work should develop stronger theoretical and empirical foundations for this style of memory-level continual learning. More broadly, RIZZ points toward compound AI systems whose persistent state is not merely an ever-growing transcript, but a curated, routed, and auditable substrate for continual adaptation.


\clearpage
\bibliographystyle{plainnat}
\bibliography{references}

\appendix

\section{RIZZ Formalism and Appendix Algorithms}
\label{app:rizz-formalism}

This appendix formalizes the mechanisms described in Section~\ref{sec:methods}: hierarchical routing, branch-local memory, retrieval, prompt compilation, and verifier-gated memory evolution. We use $S_t$ only for adaptive state and $\Gamma_t$ only for retrieved material.

\subsection{State and input stream}

At step $t$, RIZZ receives an input stream element
\begin{equation}
    z_t=(x_t,I_t,m_t,d_t,y_t^\star),
\end{equation}
where $x_t$ is the current input, $I_t$ is an optional instruction, $m_t$ is metadata, $d_t$ is optional external evidence, and $y_t^\star$ is used only by the verifier when a response is produced. If a response is required, the frozen model $M$ produces an extracted answer $\hat y_t$, and the verifier returns
\begin{equation}
    r_t=\mu_t(\hat y_t,y_t^\star,x_t,m_t)\in[0,1].
\end{equation}
If the input only provides context, no response is generated; the input is routed and stored for later retrieval.

The adaptive state is
\begin{equation}
    S_t=(\mathcal{B}_t,\mathcal{W}_t,\mathcal{H}_t),
\end{equation}
where $\mathcal{B}_t$ is the branch bank, $\mathcal{W}_t$ is bounded working memory, and $\mathcal{H}_t$ stores routing and reward diagnostics. Each branch, or zero-interference zone, has the form
\begin{equation}
    b=(\ell_b,g_b,\mathcal{E}_b,\mathcal{P}_b,\mathcal{N}_b,\mathcal{C}_b,h_b).
\end{equation}
Here $\ell_b=(f_b,a_b)$ is a hierarchical label, $g_b$ is the spawn anchor, $\mathcal{E}_b$ contains successful episodic examples, $\mathcal{P}_b$ contains procedural rules, $\mathcal{N}_b$ contains substantive failures, $\mathcal{C}_b$ contains scoped context, and $h_b$ stores summary statistics. Input embeddings are normalized:
\begin{equation}
    e_t=\frac{\phi(x_t)}{\|\phi(x_t)\|_2},
    \qquad
    \phi:\mathcal{X}\rightarrow\mathbb{S}^{d-1}.
\end{equation}
We use $\psi$ for label-string embeddings, keeping it distinct from the input embedder $\phi$.

\subsection{Routing}

For inputs that require a semantic label, the LLM judge proposes
\begin{equation}
    \tilde{\ell}_t=(\tilde f_t,\tilde a_t)
    =
    J_\omega(x_t,I_t,\mathcal{B}_t).
\end{equation}
The application component is snapped to an existing application when
\begin{equation}
    \cos(\psi(\tilde a_t),\psi(a_b))
    \geq
    \tau_{\rm snap},
    \qquad
    \tau_{\rm snap}=0.92.
\end{equation}
Let $\ell_t=(f_t,a_t)$ denote the normalized post-snap label.

If no application clears the snap threshold, RIZZ compares same-function siblings by token overlap in the application slot. Let $\mathcal{T}(a)$ denote the nonempty hyphen-delimited tokens in application label $a$. Then
\begin{equation}
    J(a_t,a_b)
    =
    \frac{
        |\mathcal{T}(a_t)\cap\mathcal{T}(a_b)|
    }{
        |\mathcal{T}(a_t)\cup\mathcal{T}(a_b)|
    },
    \qquad
    \tau_J=0.40.
\end{equation}
The routing score for a shape-compatible branch is
\begin{equation}
    s_t(b)
    =
    \begin{cases}
        1,
        &
        \ell_b=\ell_t,
        \\[3pt]
        \frac{1}{2}+\frac{1}{2}J(a_t,a_b),
        &
        f_b=f_t
        \ \text{and}\
        J(a_t,a_b)\geq\tau_J,
        \\[3pt]
        0,
        &
        \text{otherwise}.
    \end{cases}
\end{equation}
If no compatible branch receives positive score and capacity remains, RIZZ spawns a new branch:
\[
    \operatorname{Spawn}(e,\ell)
    =
    (\ell,e,\varnothing,\varnothing,\varnothing,\varnothing,h_0).
\]

When the input provides context but no question, the judge label is undefined. RIZZ instead routes by nearest spawn anchor:
\begin{equation}
    b^\star
    =
    \arg\max_{b\in\mathcal{B}_t}
    \cos(e_t,g_b).
\end{equation}
A new context branch is spawned if $\mathcal{B}_t$ is empty or if
\begin{equation}
    \max_{b\in\mathcal{B}_t}
    \cos(e_t,g_b)
    <
    \tau_{\emptyset},
    \qquad
    \tau_{\emptyset}=0.5.
\end{equation}

\begin{algorithm}[h]
\caption{Route}
\label{alg:rizz-route}
\small
\begin{algorithmic}[1]
\Require Stream element $z=(x,I,m,d,y^\star)$; embedding $e$; state $S=(\mathcal{B},\mathcal{W},\mathcal{H})$
\Ensure Branch $b$ and response flag $\chi\in\{0,1\}$
\If{$x$ provides context but no question}
    \If{$\mathcal{B}=\varnothing$}
        \State $b\gets\operatorname{Spawn}(e,\varnothing,\varnothing)$
    \Else
        \State $b^\star\gets\arg\max_{b\in\mathcal{B}}\cos(e,g_b)$
        \If{$\cos(e,g_{b^\star})<\tau_\emptyset$ and $|\mathcal{B}|<B_{\max}$}
            \State $b\gets\operatorname{Spawn}(e,\varnothing,\varnothing)$
        \Else
            \State $b\gets b^\star$
        \EndIf
    \EndIf
    \State \Return $(b,0)$
\EndIf
\State $\rho\gets\operatorname{Shape}(x,I,m)$
\State $(\tilde f,\tilde a)\gets J_\omega(x,I,\mathcal{B})$
\State Snap $\tilde f$ to an existing function if its label-embedding score exceeds $\tau_{\rm fam}$
\State Snap $\tilde a$ within that function family if its label-embedding score exceeds $\tau_{\rm snap}$
\State $\ell=(f,a)\gets$ normalized post-snap label
\State $\mathcal{D}\gets\{b\in\mathcal{B}\}$
\For{$b\in\mathcal{D}$}
    \If{$\ell_b=\ell$}
        \State $s(b)\gets 1$
    \ElsIf{$f_b=f$}
        \State $j\gets |\mathcal{T}(a_b)\cap\mathcal{T}(a)|/|\mathcal{T}(a_b)\cup\mathcal{T}(a)|$
        \If{$j\geq\tau_J$}
            \State $s(b)\gets \frac{1}{2}+\frac{1}{2}j$
        \Else
            \State $s(b)\gets 0$
        \EndIf
    \Else
        \State $s(b)\gets 0$
    \EndIf
\EndFor
\State $b^\star\gets\arg\max_{b\in\mathcal{D}}s(b)$
\If{$s(b^\star)=0$ and $|\mathcal{B}|<B_{\max}$}
    \State $b^\star\gets\operatorname{Spawn}(e,\ell)$
\EndIf
\State \Return $(b^\star,1)$
\end{algorithmic}
\end{algorithm}

\subsection{Memory and retrieval}

Each successful example in branch $b$ is stored as
\begin{equation}
    (x_i,\hat y_i,r_i,e_i,\delta_i)\in\mathcal{E}_b,
\end{equation}
where $\delta_i$ is an optional metadata scope such as a task, session, or item identifier. A procedural rule is represented as
\begin{equation}
    p_j=(u_j,\alpha_j,\beta_j,H_j,A_j)\in\mathcal{P}_b,
\end{equation}
where $u_j$ is the rule embedding, $\alpha_j$ is the antecedent, $\beta_j$ is the recommended strategy or decision, and $H_j,A_j$ are helpful and harmful counters. Its retrieval score is
\begin{equation}
    \operatorname{score}_{\rm proc}(p_j;x_t)
    =
    \cos(e_t,u_j)
    \cdot
    \operatorname{clip}
    \left(
        1+\alpha(H_j-A_j),
        m_{\min},
        m_{\max}
    \right),
\end{equation}
with $\alpha=0.30$, $m_{\min}=0.30$, and $m_{\max}=3.00$.

RIZZ decouples read and write locality. The selected branch receives future writes, but positive examples and rules may be retrieved across branches:
\begin{equation}
    \mathcal{R}^{+}_t
    =
    \operatorname{TopK}
    \left(
        \bigcup_{b\in\mathcal{B}_t}
        \left(\mathcal{P}_b\cup\mathcal{E}^{+}_b\right);
        \operatorname{score}(\cdot,e_t)
    \right).
\end{equation}
Failure memory remains branch-local:
\begin{equation}
    \mathcal{R}^{-}_t
    =
    \operatorname{TopK}
    \left(
        \mathcal{N}_{b_t};
        \cos(\cdot,e_t)
    \right).
\end{equation}

Scoped context memory is
\begin{equation}
    \mathcal{C}_b
    =
    \{(c_i,v_i,\delta_i,t_i)\}_{i=1}^{n_b^{\rm ctx}},
    \qquad
    v_i=\frac{\phi(c_i)}{\|\phi(c_i)\|_2}.
\end{equation}
If metadata provides a scope key $\delta_t$, context retrieval first filters to
\begin{equation}
    I_t=\{i:\delta_i=\delta_t\}.
\end{equation}
If this set is empty, retrieval falls back to all context stored in $\mathcal{C}_{b_t}$. The retrieved context combines semantic similarity and recency:
\begin{equation}
    \operatorname{Ctx}_{b_t}(e_t,m_t)
    =
    \operatorname{Top}_{k-r}
    \left(
        I_t;\cos(e_t,v_i)
    \right)
    \cup
    \operatorname{Recent}_{r}(I_t).
\end{equation}
The full retrieved material is
\begin{equation}
    \Gamma_t
    =
    \mathcal{R}^{+}_t
    \cup
    \mathcal{R}^{-}_t
    \cup
    \operatorname{Ctx}_{b_t}(e_t,m_t)
    \cup
    \mathcal{W}_t
    \cup
    d_t.
\end{equation}

\subsection{Prompt compilation}

The compiler maps retrieved material into a bounded prompt:
\[
    p_t=\operatorname{Compile}(x_t,I_t,\Gamma_t,B_{\rm ctx}).
\]
It follows the priority order described in Section~\ref{sec:methods}: scoped context, memory-use preamble, instruction, external evidence, procedural rules, episodic examples, common mistakes, recent working memory, and the current input. Formally, this approximates
\begin{equation}
    \max_{\mathcal{A}\subseteq\Gamma_t}
    \sum_{c\in\mathcal{A}} v(c\mid x_t,b_t)
    \quad
    \text{s.t.}
    \quad
    \sum_{c\in\mathcal{A}}\operatorname{tokens}(c)\leq B_{\rm ctx},
\end{equation}
using a fixed priority cascade rather than exact combinatorial optimization.

\subsection{Verifiers}
\label{app:ver}
Each subtask is scored with its upstream metric when available: per-task classification, ROUGE-L, SARI, fuzzy ratio, first-character accuracy, and exact-match on the eight TRACE tasks; token-F1, the official pathology-label match, and code execution on \textsc{HotpotQA-distract}, \textsc{DDXPlus},
and \textsc{DS-1000} respectively; and the LongMemEval paper's binary LLM-as-judge on LongMemEval-S. The same verifier is applied to every method on a given subtask, so per-method differences reflect retrieval and memory behavior rather than scoring. A binary LLM-judge is otherwise used.

\section{Per-sub-task Accuracy}
\label{app:per-sub-task}
\begin{table}[tbp]
\centering
\caption{\textbf{Per-sub-task accuracy} on three benchmarks (Anthropic Claude Haiku 4.5).
TRACE and StreamBench use each benchmark's native continuous-reward metric;
LongMemEval-S is reported per question category. \textbf{Bold} = best per row.
 * context-window collapse on ACE published config.
LME-S row $n$ shown in column 1 (the 143-question subset reached by Frozen/ACE
that v34\_D has not yet processed is omitted from the LME-S section).}

\label{tab:per_subtask}
\small
\setlength{\tabcolsep}{4pt}
\begin{tabular}{l l c c c c}
\toprule
& \textbf{Sub-task} ($n$) & \textbf{Frozen} & \textbf{AMem} & \textbf{ACE}\textsuperscript{*} & \textbf{RIZZ (Ours)} \\
\midrule
\multirow{8}{*}{\rotatebox{90}{\textbf{TRACE}}}
  & 20Minuten         (100) & 0.3764 & \textbf{0.3906} & 0.3851 & 0.3891 \\
  & C-STANCE          (800) & 0.7525 & 0.7200 & 0.0850 & \textbf{0.9163} \\
  & FOMC              (700) & 0.6900 & 0.7200 & 0.0000 & \textbf{0.7643} \\
  & MeetingBank       (600) & 0.1577 & 0.1912 & 0.0007 & \textbf{0.1946} \\
  & NumGLUE-cm        (243) & 0.9012 & 0.8025 & 0.0000 & \textbf{0.9465} \\
  & NumGLUE-ds        (200) & 0.6850 & 0.5150 & 0.0000 & \textbf{0.8200} \\
  & Py150             (500) & 0.5666 & \textbf{0.6019} & 0.1206 & 0.5656 \\
  & ScienceQA         (400) & \textbf{1.0000} & \textbf{1.0000} & 0.0000 & \textbf{1.0000} \\
\midrule    
\multirow{3}{*}{\rotatebox{90}{\textbf{SB}}}
  & ddxplus          (1764) & 0.6570 & \textbf{0.7120} & 0.0908 & 0.7007 \\
  & ds-1000          (1000) & \textbf{0.5360} & 0.5220 & 0.4490 & 0.4670 \\
  & hotpotqa         (1500) & 0.5793 & 0.5867 & 0.0827 & \textbf{0.6667} \\
\midrule
\multirow{4}{*}{\rotatebox{90}{\textbf{LME-S}}}
  & single-session-user      (70) & 0.3429 & 0.214 & 0.271 & \textbf{0.7429} \\
  & single-session-preference (30) & 0.3667 & \textbf{0.767} & 0.067 & 0.5333\\
  & multi-session           (133) & 0.1579 & 0.331 & 0.068 & \textbf{0.4737} \\
  & temporal-reasoning      (124) & 0.1350 & 0.271 & 0.0000 & \textbf{0.3459}\\

\bottomrule
\end{tabular}
\end{table}
\FloatBarrier

\section{ACE Failures and Attempted Workarounds}

ACE is one of the strongest published online-learning baselines and an architectural inspiration for RIZZ through its evolving playbook framework. Its failure modes therefore merit closer examination. We identify three issues in the released implementation.

\paragraph{The token budget is advisory rather than enforced.}
ACE inserts the playbook budget into the prompt as a textual instruction instead of truncating memory after updates. In practice, the model ignores the constraint. Under the published 80K setting, the playbook exceeds 130K tokens by step 290 on StreamBench. On TRACE, prompts eventually overflow Haiku’s context window entirely, producing hard failures rather than incorrect answers.

\paragraph{Per-record pruning dominates runtime.}
Once the playbook exceeds 5K tokens, ACE re-scores the entire playbook after every interaction using the \texttt{BulletpointAnalyzer}. With playbooks exceeding 100K tokens, these passes take roughly 100 seconds each on Haiku and account for most of ACE’s runtime and cost overhead.

\paragraph{The released implementation does not support conversational memory benchmarks.}
The public ACE repository contains no evaluation pipeline for benchmarks such as LongMemEval-S. In practice, this disables the Reflect and Curate stages entirely, reducing ACE to generator-only inference. We verified this from call counts in the released runs: roughly one LLM call per example instead of the expected three-stage pipeline.

\subsection{Workaround 1: ACE-tuned}
We lowered the configured playbook budget to 30K, reduced the pruning trigger, and increased the merge threshold. Results on TRACE remained effectively unchanged. This follows directly from the first issue above: because the budget is only expressed as prompt text, changing the number alters the instruction but not only slightly nudges the behavior.

\subsection{Workaround 2: ACE hard-capped}
We implemented three modifications: hard post-update truncation, a bypass for expensive whole-playbook rescoring once the cap is active, and online write-back support for LongMemEval-S. These patches keep the playbook bounded and restore end-to-end execution of the full ACE pipeline across all benchmarks. We report this variant as a fairness audit rather than a faithful reproduction, since the truncation mechanism is our addition.

We restrict claims here to mechanism-level behavior and partial LongMemEval-S results. On the first 138 LongMemEval-S instances, the subset that hard-capped ACE was able to complete before its checkpoint, it achieved an average reward of $0.10$, compared to $0.20$ for default ACE on the same instances and $0.082$ for default ACE averaged over the full set. Hard-capping therefore enables the pipeline to run to completion but does not improve per-instance reward; the aggressive truncation appears to discard bullets faster than the Reflect+Curate stages can accumulate replacements, so the playbook never reaches a useful steady state. Across both variants, recovery remained limited, consistent with the hypothesis that ACE's single mutable playbook is poorly matched to scenarios where relevant evidence is distributed across long conversational histories.

The partial runs also exposed an evaluator interaction on StreamBench DDXPlus. The published evaluator extracts the first numeric substring in the model output as the diagnosis. ACE's verbose generations often begin with numbers from patient history, causing correct answers to be mis-scored. In audited samples, ACE selected the correct diagnosis in roughly 70\% of cases while receiving only 9\% measured accuracy. Frozen, AMem, and RIZZ avoid this issue by producing tightly constrained outputs whose first numeric token is the predicted diagnosis.

\section{Extended Ablations}
\label{app:extended_ablations}


\begin{table}[tbh!]
\centering
\caption{\textbf{Threshold sensitivity on TRACE.}
Each row reports reward on $n{=}720$ examples. The default settings are $\theta{=}0.40$ for Jaccard and $\theta{=}0.92$ for snapping.}
\label{tab:threshold_sensitivity}
\small
\renewcommand{\arraystretch}{1.15}
\setlength{\tabcolsep}{10pt}
\begin{tabular}{@{}lcc@{}}
\toprule
\textbf{Threshold type} & \textbf{$\theta$} & \textbf{Reward} \\
\midrule
\multirow{5}{*}{Jaccard}
& 0.10 & 0.5055 \\
& 0.20 & 0.5211 \\
& 0.30 & 0.5435 \\
& 0.40 & \textbf{0.5481} \\
& 0.50 & 0.5468 \\
\midrule
\multirow{5}{*}{Snap}
& 0.80 & 0.5162 \\
& 0.85 & 0.5157 \\
& 0.92 & \textbf{0.5435} \\
& 0.95 & 0.5384 \\
& 0.99 & 0.5158 \\
\bottomrule
\end{tabular}
\end{table}

\begin{table}[tbh!]
\centering
\caption{\textbf{Seed robustness for RIZZ on TRACE.}
Each run uses the same configuration on $n{=}720$ examples.}
\label{tab:seed_robustness}
\small
\renewcommand{\arraystretch}{1.05}
\setlength{\tabcolsep}{6pt}
\begin{tabular}{@{}l r@{}}
\toprule
\textbf{Run} & \textbf{Reward} \\
\midrule
Seed 0 & 0.5435 \\
Seed 1 & 0.5482 \\
Seed 2 & 0.5332 \\
\midrule
Mean $\pm$ std & 0.5416 $\pm$ 0.0077 \\
Range & 0.0150 \\
\bottomrule
\end{tabular}
\end{table}

\begin{table}[tbh!]
\centering
\caption{\textbf{Ablation on TRACE} for $n{=}720$ instances. Deltas are percentage-point changes relative to baseline RIZZ.}
\label{tab:trace_ablation_merge_prune}
\small
\renewcommand{\arraystretch}{1.15}
\setlength{\tabcolsep}{8pt}
\begin{tabular}{@{}lc@{}}
\toprule
\textbf{Run} & \textbf{Reward} \\
\midrule
\texttt{- merge\_sweep} & 0.5204 \dn{2.12} \\
\texttt{- prune}        & 0.5216 \dn{2.00} \\
\bottomrule
\end{tabular}
\end{table}

\begin{table}[t]
\centering
\caption{\textbf{Router judge ablation on TRACE.}
Average reward over $n{=}720$ examples.}
\label{tab:router_judge_ablation}
\small
\renewcommand{\arraystretch}{1.15}
\setlength{\tabcolsep}{8pt}
\begin{tabular}{@{}l c@{}}
\toprule
\textbf{Router judge} & \textbf{Avg.\ reward} \\
\midrule
GPT-4.1 & 0.5224 \\
\rowcolor{ourrow}
\textbf{Claude Haiku 4.5} & \textbf{0.5929} \\
\bottomrule
\end{tabular}
\end{table}

\FloatBarrier

\section{Extended Related Work}
\label{app:relatedwork}

RIZZ lies at the intersection of black-box language-model adaptation, reflective prompt optimization, test-time memory, agentic context engineering, and long-term memory for LLM agents. The common theme across these literatures is that modern language systems can improve without directly updating the parameters of the base model. The critical distinction is \emph{what object is adapted}. Prompt optimizers adapt instructions; context-engineering methods adapt playbooks; memory systems adapt retrieval state; agentic memory systems adapt graph-structured historical knowledge; and continual-learning benchmarks measure whether this adaptation persists without catastrophic interference. RIZZ is closest to the first four categories, but differs in treating continual adaptation as a verifier-gated, branch-structured routing and memory problem rather than as optimization of a single prompt, single context, or single global memory.

\subsection{Black-box adaptation through natural-language feedback}

A growing body of work replaces gradient or reward-only learning with natural-language feedback. Reflexion \cite{shinn2023reflexionlanguageagentsverbal} showed that language agents can improve across trials by verbally reflecting on feedback and storing those reflections in episodic memory, avoiding weight updates while improving decision-making, coding, and reasoning. Self-Refine \cite{selfrefine} similarly demonstrated that a model can generate an initial output, critique it, and iteratively refine it using the same model as generator, feedback provider, and refiner. These methods established the viability of test-time textual self-improvement, but they typically operate within a single episode or maintain relatively simple reflective memory.

TextGrad \cite{yuksekgonul2024textgradautomaticdifferentiationtext} generalizes this idea into a PyTorch-like framework for automatic ``differentiation'' through text: LLM-generated feedback is backpropagated through textual variables in a computation graph to improve prompts, code, molecules, or other textual artifacts. GEPA \cite{GEPAmain} pushes this direction toward rollout-efficient optimization of compound AI systems. Given a system with one or more prompts, GEPA samples trajectories, reflects on failures in natural language, proposes prompt mutations, and combines complementary lessons from a Pareto frontier of candidates. Feedback Descent \cite{lee2025feedbackdescentopenendedtext} extends the same principle to open-ended text-artifact optimization by preserving pairwise textual critiques as high-bandwidth directional information rather than collapsing comparisons to scalar preferences.

RIZZ shares with these systems the premise that language is a useful optimization medium. However, the update target is different. GEPA and Feedback Descent optimize one or more textual artifacts over an evaluation distribution. RIZZ instead maintains a persistent, heterogeneous external state that must route each new instance to the right memory branch, retrieve relevant experience, compile a bounded prompt, and update only after verifier feedback. Thus, the key question is not only ``how should a prompt change?'' but also ``which accumulated experience should condition this input, and when is a new experience safe to remember?''

\subsection{Test-time memory and Dynamic Cheatsheet}

Dynamic Cheatsheet \cite{suzgun2025dynamiccheatsheettesttimelearning} is one of the closest precursors to RIZZ. It introduces a lightweight framework that gives a black-box language model a persistent, evolving memory at inference time. Instead of solving each query independently, the model stores and reuses accumulated strategies, code snippets, and problem-solving insights; the system uses a generator to solve the current query and a curator to update the memory with useful, compact lessons. Dynamic Cheatsheet is explicitly non-parametric and black-box: it adapts by changing memory rather than model weights.

RIZZ adopts the same broad philosophy of adaptive test-time memory but addresses limitations that arise in heterogeneous continual streams. Dynamic Cheatsheet largely assumes a single evolving memory, optionally augmented by retrieval and synthesis. RIZZ decomposes memory into dynamically spawned branches, each with procedural rules, episodic examples, reward statistics, a subspace sketch, and a behavioral summary. This matters when the stream contains multiple latent task families: a Game-of-24 code heuristic, a medical-diagnosis heuristic, and a multimodal calibration heuristic should not necessarily coexist in one undifferentiated cheatsheet. RIZZ therefore turns memory selection into an explicit routing problem and uses verifier feedback to decide not only what to write, but also where to write it.

The second difference is write discipline. Dynamic Cheatsheet can self-curate without ground-truth labels, which is powerful in unlabeled settings but risks self-reinforcing incorrect abstractions when the model's own judgment is unreliable. RIZZ\ instead uses deterministic verifiers whenever available and treats verification as the gate for durable memory. High-reward interactions can be promoted into rules or exemplars; low-reward interactions can become anti-patterns; and rules are promoted only after repeated verified usefulness. The resulting system is more conservative, but better suited to benchmarked continual adaptation where exact match, F1, nDCG, code execution, trajectory success, or scientific-plot scores are available.

\subsection{Agentic Context Engineering and evolving playbooks}

Agentic Context Engineering (ACE) \cite{zhang2026agenticcontextengineeringevolving} generalizes Dynamic Cheatsheet from compact memory snippets to comprehensive evolving playbooks. ACE frames context adaptation as modifying the input context---system instructions, strategies, evidence, memory, or domain-specific guidance---rather than changing model weights. It introduces a modular generation--reflection--curation workflow that accumulates, refines, and organizes strategies while avoiding brevity bias and context collapse, where iterative rewriting erodes useful domain detail. ACE's central lesson is that context should not merely become shorter; it should become better organized.

RIZZ follows this lesson but applies it at a finer granularity. Instead of maintaining one evolving playbook for the whole agent or task domain, RIZZ maintains branch-local playbooks, family-level playbooks, and global playbooks. This design is motivated by continual-learning interference: a rule that is valuable for one task family may be harmful for another, and broad promotion should require evidence across branches. ACE asks how to evolve context so an agent becomes stronger over repeated experience. RIZZ asks how to evolve many contexts simultaneously, how to route a new instance to the right one, how to spawn new contexts when the stream changes, and how to prevent unverified or branch-specific heuristics from contaminating the global context.

\subsection{Agentic long-term memory and A-Mem}

A separate line of work studies long-term memory systems for LLM agents. Retrieval-Augmented Generation \cite{lewis2021retrievalaugmentedgenerationknowledgeintensivenlp} combines parametric and non-parametric memory by retrieving external passages and conditioning generation on them. MemGPT \cite{packer2024memgptllmsoperatingsystems} introduces virtual context management inspired by operating-system memory hierarchies, allowing an LLM to move information between memory tiers to support long documents and multi-session interaction. These systems improve access to external information, but do not by themselves define a policy for experience curation, memory evolution, or cross-task interference control.

A-Mem \cite{xu2025amemagenticmemoryllm} advances this area by making the memory system itself more agentic. Inspired by the Zettelkasten method, A-Mem constructs structured memory notes with contextual descriptions, keywords, tags, and links to related memories. Adding a new memory can trigger link generation and evolution of existing memory attributes, allowing the memory network to reorganize as higher-order patterns emerge. A-Mem improves the internal organization of memory objects, especially through semantic links and evolving notes.

RIZZ is complementary to A-Mem. A-Mem helps retrieve semantically connected facts or experiences. Our method controls allocation of experience across branches and routing of queries to memory namespaces. The distinction is crucial: two inputs may be semantically close yet procedurally incompatible because they require different answer formats, reasoning styles, tools, or verifiers. RIZZ's branch-specific subspace sketches, contextual-bandit routing, output-shape features, and reward histories directly address this incompatibility. In a combined system, A-Mem-style semantic linking can serve as a retrieval substrate inside RIZZ, while RIZZ provides the higher-level continual-learning policy: when to create a new branch, which branch to route to, which memories to expose, which rules to promote, and which failures to store as anti-patterns. A-Mem is also dependent heavily on an embedder. For example, on performing an ablation on different embedders on a LongMemEval subset of 147 samples specifically on AMem. We observe that reward is $0.231$ for MiniLM and $0.109$ for Qwen3VL-2B.

\subsection{Continual learning benchmarks and interference control}

The broader lifelong-agent literature emphasizes that LLM agents deployed in dynamic environments should accumulate, transfer, and retain knowledge over time. StreamBench \cite{wu2024streambenchbenchmarkingcontinuousimprovement} evaluates whether language agents can improve over an input--feedback sequence. TRACE \cite{wang2023tracecomprehensivebenchmarkcontinual} evaluates continual learning across heterogeneous tasks and measures forgetting and transfer under sequential exposure. LongMemEval \cite{wu2025longmemevalbenchmarkingchatassistants} probes multi-session long-term memory in chat assistants.

These benchmarks foreground the stability--plasticity dilemma \cite{french1999catastrophic, Kirkpatrick_2017, qiang2026plasticitystabilityposttraininglarge}. A system must be plastic enough to learn new task families, but stable enough not to corrupt previously useful behavior. Many LLM memory systems improve plasticity by writing new memories, yet do not isolate incompatible experiences. RIZZ addresses this through branch-local memory, novelty-triggered branch creation, learned routing, branch merge/prune lifecycle control, and tiered promotion from branch to family to global memory. This resembles mixture-of-experts specialization at the memory level: branches act as non-parametric experts, and the router learns which expert should condition each input.

\section{Benchmarks on Quantum Calibration Data}
\label{app:qcaleval}

We include QCalEval~\cite{cao2026qcalevalbenchmarkingvisionlanguagemodels} as a stress test on online test stream~\cite{shen2024rethinking, schorling2025meta, vaidhyanathan2025metasym}.
QCalEval evaluates vision-language models on quantum calibration plots, with 243 test
instances and six questions per instance covering technical plot description, experimental
classification, significance assessment, fit reliability, parameter extraction, and calibration
diagnosis. This benchmark is useful for RIZZ because it combines multimodal evidence,
sparse verifier feedback, and strict output-format constraints. 

We evaluate the public QCalEval over full sweeps, giving
$243 \times 6 = 1458$ model calls per pass. The frozen actor is Claude Opus 4.6. Final reporting uses the official QCalEval judge pipeline: free-form questions Q1 and Q3 are judged by the benchmark LLM judge, Q5 is scored by the parameter-extraction
metric, and Q2, Q4, and Q6 are scored as classification questions.

In Pass 1, each query is presented as the task instruction, image
content, and question prompt, with no retrieved demonstrations. The resulting trajectories
are verifier-scored, and successful Q3/Q5/Q6 episodes are written to our branch bank. All questions are tested in further passes. No model
weights are updated in any pass, and test-set gold answers are never included in prompts.

\begin{table*}[tbh!]
\centering
\small
\caption{\textbf{QCalEval supplementary results.}
Scores are reported as percentages, over 243 test entries and six questions per entry.
RIZZ uses a frozen Claude Opus 4.6 actor in the passes.}
\label{tab:qcaleval}
\renewcommand{\arraystretch}{1.15}
\setlength{\tabcolsep}{8pt}
\begin{tabular}{@{}l r@{}}
\toprule
\textbf{Model} & \textbf{Overall} \\
\midrule
Claude Opus 4.6 & 67.8 \\
Ising-Cal-1 & 74.7 \\
\rowcolor{ourrow}
\textbf{Claude Opus 4.6 + RIZZ} & \textbf{80.4} \\
\bottomrule
\end{tabular}
\end{table*}

\section{Compute Resources}

All runs were performed on Modal. The Ising-Cal-1 model was hosted on NVIDIA 2xH100 GPUs with 80GB of GPU memory. The Qwen-3VL-2B embedder was hosted on NVIDIA L4 GPU. These runs cost 2000 USD on Modal. Anthropic API costs were around 2500 USD and OpenAI costs were around 500 USD. 



\clearpage

\end{document}